\documentclass{article}

\usepackage{graphicx} 
\usepackage{subfigure}
\usepackage{graphicx}
\usepackage{amsmath}
\usepackage{algorithm}
\usepackage{algorithmic}

\usepackage{url,enumerate}
\usepackage{color}
\usepackage{hyperref}
\usepackage{natbib}
\renewcommand{\cite}[1]{\citep{#1}}

\input{Definitions}

\newcommand{\commenttext}[1]{}
\newcommand{\emcite}[1]{\citet{#1}}
\newcommand{\np}[1]{\nabla\psi(#1)}

\newcommand{\rk}{\textrm{rk}}

\begin{document}

\title{Distributed Autonomous Online Learning: Regrets and Intrinsic Privacy-Preserving Properties}

\author{Feng Yan \\
        Department of CS\\ Purdue University \\
        \and
        Shreyas Sundaram \\
        Department of ECE\\ University of Waterloo \\
        \and
        S.V\!.\,N. Vishwanathan
        \\
        Departments of Statistics and CS\\ Purdue University \\
        \and
        Yuan Qi \\
        Departments of CS and Statistics \\
        Purdue University
}
\maketitle


%

\begin{abstract}
Online learning has become increasingly popular on handling massive data. The sequential nature of online learning, however, requires a centralized learner to store data and update parameters. In this paper, we consider online learning with {\em distributed} data sources. The autonomous learners update local parameters based on local data sources and periodically exchange information with a small subset of neighbors in a communication network.
We derive the regret bound for strongly convex functions that generalizes the work by \citet{RamNedVee09} for convex functions. More importantly, we show that our algorithm has \emph{intrinsic} privacy-preserving properties, and we prove the sufficient and necessary conditions for privacy preservation in the network. These conditions imply that for networks with greater-than-one connectivity, a malicious learner cannot reconstruct the subgradients (and sensitive raw data) of other learners, which makes our algorithm appealing in privacy sensitive applications.
\end{abstract}

\section{Introduction}
\label{sec:Introduction}

Online learning has emerged as an attractive paradigm in
machine learning given the ever-increasing amounts of data being
collected everyday. It efficiently reduces the training time by processing the
data only once, assuming that all the training data are available at a central
location. For many applications, however, this assumption is problematic. For instance,
sensor networks may be deployed in rain forests and collect data
autonomously. The cost of transmitting all the data to a {\em central} server can be prohibitively high.
Also, sharing sensitive data might lead to information leakage and raise privacy concerns. For example,
banks collect credit information about their customers but might not share the data with other financial institutions for privacy concerns.
Similarly privacy concerns might prevent sharing of patient records across hospitals.

Therefore it is desirable to conduct distributed learning in a {\em fully decentralized} setting. Specifically, we treat individual computational units (e.g., processors) in a network as {\em autonomous} learner. They learn model parameters independently from their local data sources, and pass estimation information to their neighbors in a communication network. By doing so, distributed learning avoids sharing original, sensitive data with others and storing data in a central location.


In this paper, we consider a general \emph{distributed autonomous online learning} algorithm to learn from {\em fully decentralized} data sources.
We address two important questions associated with this general algorithm. The first question is how the {\em distributed} online learners perform compared with the optimal learner chosen in hindsight. To this end we derive the regret bound for strongly convex functions. Our work is closely related to the recent work by \citet{RamNedVee09,NedOzd09}; the main difference lies in our analysis for {\em strongly} convex functions, which naturally extends the results of \cite{RamNedVee09}.

The second question is how the topology of the computational network affects privacy preservation. To answer this question, we draw ideas from the modern control theory to model the distributed online learning algorithm as a structured linear time-invariant system, and we establish theorems on necessary and sufficient conditions that a malicious learner can reconstruct the subgradients for other learners at other locations. Based on these conditions, we conclude that for most communication topologies, namely with connectivity greater than one, our algorithm inherently prevents the reconstruction of the subgradients at other locations, therefore avoiding information leakage. Unlike previous works on privacy-preserving learning that mostly alter the original learning algorithms by patching cryptographical tools, such as secure multi-party computation \cite{SakAra10,KeaTanWor07} and randomization \cite{ChaMon09}, or data aggregation \cite{Rup10,AviBut07}, our privacy-preserving properties are \emph{intrinsic} in the sense that they do not require any modifications to the algorithm but are solely determined by the communication network topology of the distributed learners.


The main contributions of this paper include:
\begin{itemize}
\item We present a distributed autonomous online learning algorithm that computes local subgradients and shares parameter vectors between nodes in a communication network. We derive its regret bounds for strongly convex (hence convex) functions.
\item We use results from the modern control theory to show the connection between the reconstructability of local subgradients and the topology of the communication network, which implies privacy preservation of local data for well-chosen communication networks.
\end{itemize}

\section{Preliminaries}
\label{sec:Preliminaries}

\textbf{Notation:} Lower case letters (\eg, $w$) denote (column) vectors
while upper case letters (\eg, $A$) denote matrices. We will denote the
$(j,i)$-th element of $A$ by $A_{ji}$ and the $i$-th column of $A$ by
$A_{i}$. Subscripts with $t$, $t+1$ etc are used for indexing the
parameter vector with respect to time while superscripts are used for
indexing with respect to a processor. For instance, $w_{t}^{i}$ denotes
the parameter vector of the $i$-th processor at time $t$. We use $e_i$
to denote the $i$-th basis vector (the vector of all zeros except one on
the $i^{\mathrm{th}}$ position), and $e$ to denote the vector of all
ones. Unless specified otherwise, $\nbr{\cdot}$ refers to the Euclidean
norm $\nbr{x} := \left(\sum_{i} x_{i}^{2}\right)^ {1/2}$, and
$\inner{\cdot}{\cdot}$ denotes the Euclidean dot product $\inner{x}{x'}
= \sum_{i} x_{i} x'_i$.

\textbf{Sequential Online Learning:} Online learning usually proceeds in
trials. At each trial a data point $x_{t}$ is given to the learner which
produces a parameter vector $w_{t}$ from a convex set $\Omega \subseteq
\RR^{n}$. One then computes some function of the inner product
$\inner{w_{t}}{x_{t}}$ in order to produce a label $\hat{y}_{t}$. The
true label $y_{t}$ is revealed to the learner, which then incurs a
convex (but not necessarily smooth) loss $l(w_{t}, x_{t}, y_{t})$ and
the learner adjusts its parameter vector. If we succinctly denote $f_{t}(w) := l(w,
x_{t}, y_{t})$, then online learning is equivalent to solving the
following optimization problem in a stochastic fashion:
\begin{align}
  \label{eq:stoc-obj-serial}
  \min_{w \in \Omega} J(w), \text{ where } J(w) = \sum_{t=1}^{T}
  f_{t}(w) \text{ and } \Omega \subseteq \RR^{n},
\end{align}
and the goal is to minimize the regret
\begin{align}
  \label{eq:regret-serial}
  \mathcal{R}_S = \sum_{t=1}^{T} f_{t}(w_{t}) - \min_{w \in \Omega} J(w).
\end{align}

For many applications, however, the data are not all available to a centralized learner to
perform sequential online learning.

\textbf{Communication via Doubly Stochastic Matrix: }
We shall see that our
autonomous learners exchange information with their neighbors. The
communication pattern is defined by a weighted directed graph with a $m$-by-$m$
adjacency matrix, $A$, is doubly stochastic. Recall that a matrix is
said to be doubly stochastic if and only if all elements of $A$ are
non-negative and both rows and columns sum to one.
%
%

In the following analysis of regret bounds, we are interested in the limiting behaviors of $A^k$ as $k\rightarrow\infty$. It is well known in finite-state Markov chain theory that there are geometric bounds for $A^k$ if $A$ is \emph{irreducible} and \emph{aperiodic}~\cite{Liu01}:
\begin{align}
    \label{eq:markov-def}
    \forall i, \; & \sum_{j} |A_{ji}^k - 1/m| \leq C \beta^{k}, \\
    & C>0 \textrm{ and } 0<\beta<1. \nonumber
\end{align}
where $C$ and $\beta$ depend on the size and the topology of $G$. For example, the famous spectral geometric bound has $C=\sqrt{m}, \beta=\textrm{the spectral gap of }A$. To this end, \citet{DucAgaWai10} examined the impact of different choices of $A$ and network topologies on the convergence rate of the dual averaging algorithm for distributed optimization. Since the relationship between network topology and convergence rate is not the focus of this paper, we use the bound given in Chapter 12 of \cite{Liu01} in this paper for simplicity, where $C=2$ and $\beta$ is related to the minimum non-zero values of $A$. It is easy to show that our regret bounds can be modified accordingly if one use a general Markov mixing bound.

\section{Distributed Autonomous Online Learning}
\label{sec:Coopautononline}

For distributed autonomous online learning, we
assume to have $m$ local online learners using only data stored at
local sites.  At each trial $m$ data points $x_{t}^{i}$ with $i \in
\cbr{1, 2, \ldots, m}$ are given and the $i$-th learner updates model
parameters based on the $i$-th point.  The learner produces a parameter
vector $w_{t}^{i}$ which is used to compute the prediction
$\inner{w_{t}^{i}}{x_{t}^{i}}$ and the corresponding loss $f_{t}^{i}(w)
= l(w, x_{t}^{i}, y_{t}^{i})$. The learners then {\em exchange}
information with a selected set of their neighbors before updating
$w_{t}^{i}$ to $w^{i}_{t+1}$.  The
communication pattern amongst processors is assumed to form a strongly
(but not necessarily fully) connected graph. In particular, we will
assume a directed weighted graph whose adjacency matrix $A$ is doubly
stochastic. One can interpret the entry $A_{ji}$ as the importance that
learner $i$ places on the parameter vector communicated by learner
$j$. Of course, if $A_{ji} = 0$ then learners $j$ does not send data to learner $i$.

The corresponding optimization problem is
\begin{align}
  \label{eq:stoc-obj}
  \min_{w \in \Omega} J(w) = \sum_{t=1}^{T}
  \sum_{i=1}^{m} f_{t}^{i}(w) \text{ and } \Omega \subseteq \RR^{n},
\end{align}
and regret is measured with respect to the parameter vector $w_{t}^j$ of an \emph{arbitrary} learner $j$:
\begin{align}
  \label{eq:regret-dist}
  &\mathcal{R}_{DA} = \sum_{t=1}^{T} \sum_{i=1}^{m} f_{t}^{i}(w_{t}^j) -
  \min_{w \in \Omega} J(w)
\end{align}

If we denote $f_t=\sum_{i=1}^m f_t^i$\footnote{We abuse the notation
  $f_t$ hereinafter.}, our definition of the regret has the same form of
the regret in sequential online learning for each local learner. Given $N$ data points, there are $T = N$ iterations or trial in sequential online learning. In our case, this number reduces down to $T = \frac{N}{m}$.


We will show the convergence of $w_t^j$ by bounding the regret
$\mathcal{R}_{DA}$. In particular, we are interested in generalizing the
celebrated $\sqrt{T}$ and $\log{T}$
bounds~\cite{Zinkevich03,HazAgaKal07} of sequential online learning to
distributed autonomous online learning.

We present a general online learning algorithm for solving
\eqref{eq:stoc-obj} here. Specifically, a local learner propagates the
parameter to other learners. After receiving the parameters from other
learners, each learner updates its local parameter through a linear
combination of the received and its own old parameter. Then the local
learner updates the local model parameter based on the data collected
and the local subgradient. Via this cooperation, the learners learn a
model from distributed data sequentially. The algorithm is summarized in
Algorithm \ref{alg:stoc-grad-desc}.

\begin{algorithm}[t]
  \caption{Distributed Autonomous Online Learning}
  \label{alg:stoc-grad-desc}
  \begin{algorithmic}[1]
    \STATE \textbf{Input:} The number of learners $m$; initial points
    $w_{1}^{1}, \ldots w_{1}^{m}$; double stochastic matrix $A=(A_{ji}) \in
    \RR^{m \times m}$; and maximum iterations $T$.
    \FOR{$t=1, \ldots, T$}
    \FOR{each learner $i=1, \ldots, m$}

    \STATE $g_{t}^{i} \leftarrow \partial_{w} f_{t}^{i}(w_{t}^{i})$.

    \STATE Communicate $w_{t}^{i}$ with neighbors (as defined by $A$)
    and obtain their parameters.

    \STATE $\hat{w}_{t+1}^{i} \leftarrow \sum_{j} A_{ji} w_{t}^{j} - \eta_t g_{t}^{i}$ (Local subgradient descent) \label{line:local-subgradient}

    \STATE \small $w_{t+1}^i \leftarrow P_{\Omega} \rbr{\hat{w}_{t+1}^i} = \argmin_{w \in \Omega} \nbr{w - \hat{w}_{t+1}^i}$. (Projection)
    \ENDFOR


    \ENDFOR
  \end{algorithmic}
\end{algorithm}

\subsection{Regret Bounds}

\label{sec:regbound}

For our analysis we make the following standard assumptions, which are
assumed to hold for all the proofs and theorems presented below.
1) Each $f_{t}^{i}$ is strongly convex with modulus $\lambda \geq
  0$\footnote{Note that we allow for $\lambda = 0$, in which case
    $f_{t}^{i}$ is just convex, but not strongly convex.}.
2) $A_{ji} \neq 0$ if and only if the $i^{\mathrm{th}}$ learner
  communicates with the $j^{\mathrm{th}}$ learner. We further assume $A$
  is irreducible, aperiodic, and there exists $\beta<1$ as defined in \eqref{eq:markov-def}.
3) $\Omega$ is a closed convex subset of $\RR^{n}$ with non-empty
  interior. The subgradient $\partial_{w} f_{t}^{i}(w)$ can be computed for
  every $w \in \Omega$.
4) The diameter $\text{diam}(\Omega) = \sup_{x,x' \in \Omega} \nbr{x-x'}$ of $\Omega$ is bounded by $F < \infty$.
5) The set of optimal solutions of \eqref{eq:stoc-obj} denoted by
  $\Omega^{*}$ is non-empty.
6) The norm of the subgradients of $f_t^i$ is bounded by $L$, and $w_1^i$ are identically initialized.

The following theorem characterizes the regret of Algorithm \ref{alg:stoc-grad-desc}. The proof can be found in the appendix.

\begin{theorem}
\label{thm:reg-bound}
  If $\lambda > 0$ and we set $\eta_{t} = \frac{1}{2 \lambda t}$ then
  \begin{align}
  \label{eq:log-bound}
    \sum_{t=1}^{T} f_{t}(w_{t}^j) - f_{t}(w^{*}) \leq \frac{2C L^{2} m}{
      \lambda} (1 + \log(T)),
  \end{align}
  On the other hand, when $\lambda = 0$, if we set $\eta_{t} =
  \frac{1}{2 \sqrt{t}}$ then
  \begin{align}
  \label{eq:sqrt-bound}
    \sum_{t=1}^{T} f_{t}(w_{t}^j) - f_{t}(w^{*}) \leq m \rbr{F +
      4CL^{2}} \sqrt{T}.
  \end{align}
  $C=\frac{5-\beta}{1-\beta}$ is a communication-graph-dependent constant.
\end{theorem}


When $m = 1$, Algorithm \ref{alg:stoc-grad-desc} reduces to the
classical sequential online learning. Accordingly, our bounds
\eqref{eq:sqrt-bound} and \eqref{eq:log-bound} become the classical
square root regret $O(\sqrt{N})$ of \cite{Zinkevich03} and the logarithmic regret $O(\log T)$ of \cite{HazAgaKal07}. When $m > 1$, recall that for every time $t$, the $m$
processors simultaneously process $m$ data points. Therefore in $T$
steps our learners process $mT$ data points. If we let $N = mT$, then
our bounds can be rewritten as $O(\sqrt{mN})$ and $O(m + m \log(N/m))$,
respectively. It must be borne in
mind that our algorithm is affected by two limiting factors. First,
there is only limited information sharing between different learners.
Second, by our definition of regret, our algorithm is forced to predict
on $m$ data points in one shot with a single parameter vector
$w_{t}^j$. This is in contrast with the sequential online learner which
has access to the full data set and can use different parameter vectors
for each of the $m$ data points.

%

If we treat all the distributed parameters across the learners as a single aggregated parameter $\overline{w}_t=(w_t^1,\ldots,w_t^m)$, we can apply the results for sequential online learning to obtain the generalization bounds for distributed online learning in terms of the regret bounds. Due to space limitation, we present the generalization bounds in the appendix.

\section{Privacy and Topology of Communication Graphs}

A common form of $f_t^i(w)$ in the cost function \eqref{eq:stoc-obj}
is $l(y_t^i,\inner{w}{x_t^i})$. So the subgradient \wrt to $w_t^i$ is
$g_t^i = \partial_z l(y_t^i,\inner{w_t^i}{x_t^i}) ~x_t^i$, which is
proportional to $x_t^i$. Thus algorithms that transmit subgradients (\eg~
the first variant of Langford \ea's algorithm \cite{LanSmoZin09}) may
disclose sensitive information about raw data  (\eg, medical record), which
is undesirable for the privacy-sensitive applications mentioned, such as mining
patient information across hospitals.
Our decentralized algorithm transmits only local model parameters between neighbors in the network,
reducing the possibility of information leakage.



\begin{figure}
\begin{center}
\includegraphics[scale=0.55]{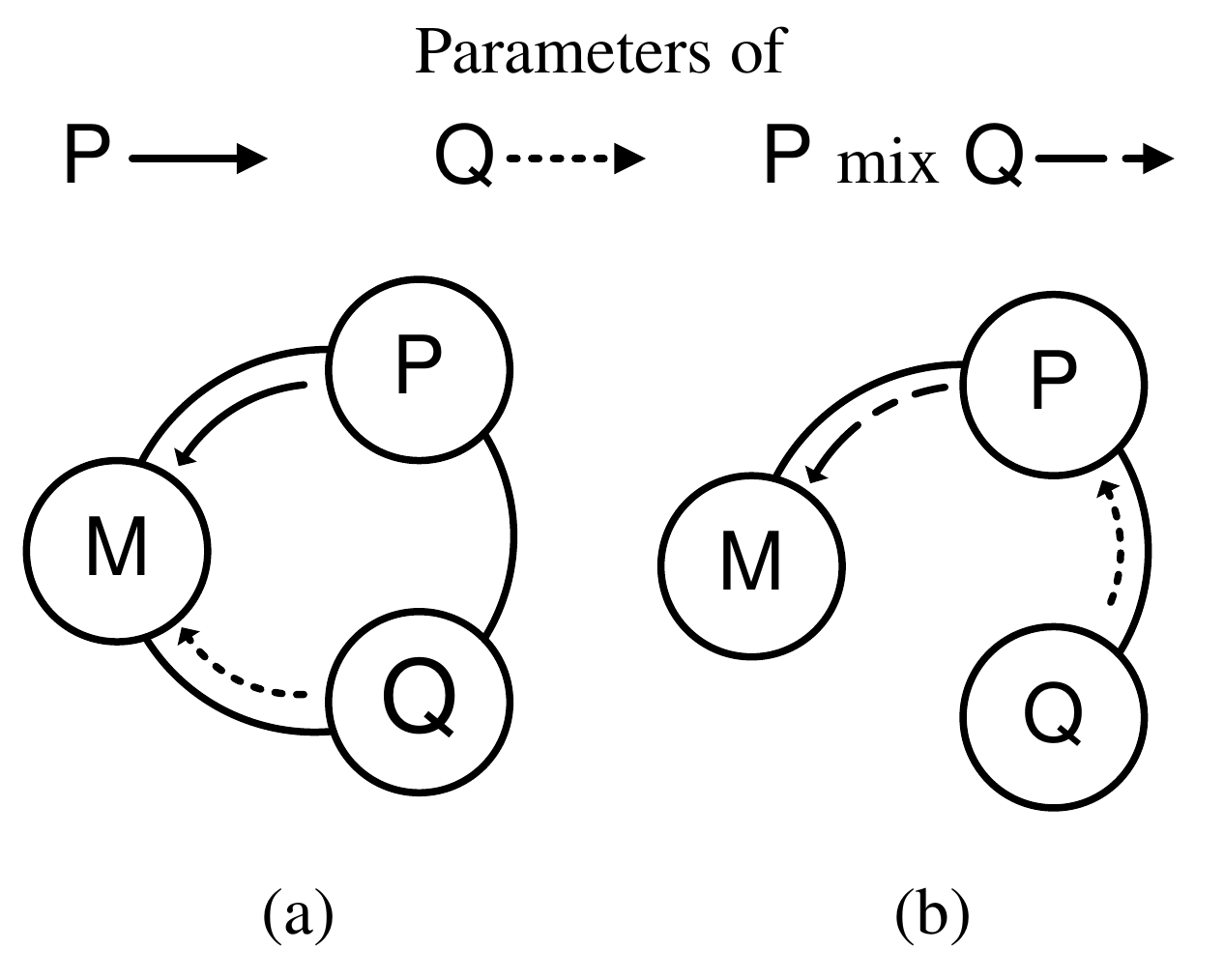}
\end{center}
\caption{Illustrating the impact of network topology on privacy preservation. In each of the three-node networks, $M$ is a malicious node (learner) that wants to gather the subgradients of $P$ and $Q$. (a) $M$ can easily reconstruct the subgradients of $P$ and $Q$ by differentiating successive parameters received from $P$ and $Q$. (b) $M$ cannot reconstruct the subgradients of $P$ and $Q$. Intuitively this is because $M$ does not receive any information from $Q$ and the parameters of $P$ is ``mixed" with $Q$'s parameters and subgradients.}
\label{fig:privacy}
\end{figure}

Formally, the communication graph is a directed graph $C(A)$. The node set consists of the online learners $\{1,\ldots,m\}$. The edge set $\Ecal$ is $\{(i,j)|A_{ij} \neq 0\}$, where node $i$ is connected to node $j$ if the weight $A_{ij}$ is nonzero. We say a node $i$ is {connected to} $j$ \emph{if and only if} $(i,j)\in\Ecal$. The neighbor set $N(j)$ of $j$ is $\{i|(i,j)\in\Ecal\}$. Intuitively the topology of the communication graph can affect the privacy-preserving capability. Consider the two examples in figure \ref{fig:privacy} to gain intuition. We assume that all nodes (learners) $M$, $P$ and $Q$ know the matrix $A$ representing the communication graph, and the convex set $\Omega = \RR^n$. Suppose $M$ is a malicious node that wants to gain information about the input data of $P$ and $Q$ by recovering their subgradients. Based on the communication graph in Figure \ref{fig:privacy}.(a), $M$ receives the parameters from $P$ and $Q$. It can use the received parameters to compute the linear combination and find the subgradient. By contrast, it is intuitively difficult to recover the subgradients based on the communication graph in Figure \ref{fig:privacy}.(b). Here $P$'s parameters are ``mixed" with the $Q$'s parameters through a linear combination at the local subgradient step (line \ref{line:local-subgradient} in Algorithm \ref{alg:stoc-grad-desc}) before sent to $M$, and $M$ does not directly receive any information from $Q$. The ambiguity about the parameters of $Q$ prevents the malicious node $M$ from correctly reconstructing the local subgradients of $P$ and $Q$.

\subsection{Full Reconstruction}

Inspired by these two examples, we formally examine under which conditions a malicious node cannot reconstruct \emph{all} subgradients of other nodes based on the parameter vectors of its adjacent nodes. We refer to this problem as \emph{full reconstruction} of subgradient, in contrast to the \emph{partial} reconstruction of subgradients discussed later. We assume $\Omega=\RR^n$ for this moment, {\em i.e.}, there is no projection step in Algorithm \ref{alg:stoc-grad-desc}. Projection will be handled differently later. Throughout this section, we shall use the following definitions and notations.
\begin{align*}
    & W_t=[{w_t^1},\ldots,{w_t^m}],~~~~G_t=[g_t^1,\ldots,g_t^m]
\end{align*}
We also assume that every learner (node) knows the whole communication matrix $A$ and the initial parameter values $W_1$ of all other learners. Without loss of generality, we may also assume the dimension of each $w_t^i$ (thus $g_t^i$) is $1$, since $W_t$ can be reconstructed row-by-row.

Now we formulate the problem of reconstructing all subgradients of the other nodes based on the following linear time-invariant dynamic system\footnote{Standard control notation is to treat the state of the system as a column vector, so that systems are written as $w_{t+1} = Aw_{t} + \tilde{G}_t$, but the state vectors in this paper are written as row vectors in order to maintain consistency with the rest of the paper.}:
\begin{align}
\label{eq:LTI-representation}
\Scal:
\begin{cases}
W_{t+1} = W_{t} A + \Gtilde_{t} \\
Y_{t} = W_{t} C
\end{cases}
\end{align}
where  $\Gtilde_t=-\eta_t G_t$ is the (unknown) \textit{input} ({\em i.e.}, local subgradients), $W_{t}$ is the \textit{state}, and $Y_t$ is the \textit{output} ({\em i.e.}, the columns of $Y_t$ are parameter vectors received by $M$), and $C$ is a matrix selecting the columns of $W_{t}$ that node $M$ receives.
According to \emcite{Bro91},  the system $\Scal$ is \textit{invertible}, if the output sequence $Y_t$ determines the unique input $\Gtilde_t$. Therefore, we can
rephrase the full subgradient reconstruction problem as the invertibility of $\Scal$.
Our theorem relates the invertibility of $\Scal$ to the topological properties of the communication graph.
\begin{theorem}
\label{thm:privacy}
If all other nodes are connected to $M$, then for almost any choice of nonzero entries in $A$, the output sequence $Y_t$ at the malicious node $M$ gives rise to a unique sequence of subgradients $\tilde{G}_t$.  On the other hand, if all other nodes are not connected to $M$, then regardless of the choice of nonzero entries in $A$, the output sequence $Y_t$ does not uniquely specify $\tilde{G}_t$.
\end{theorem}

If all other nodes are connected to M, the malicious node can reconstruct $\Gtilde_t$ by duplicating the linear combination steps at the other nodes and differentiating the successive parameter vectors. This is exactly what happens in figure \ref{fig:privacy}(a). The proof for the latter part of the theorem relies on the analysis of the generic rank of \textit{structured} systems \citep{SunHad09,DioComWou03}, which relates the rank of the transfer matrix $(zI-A)^{-1}C,~z\in\CC$ of $\Scal$ to the topological features defined by vertex disjoint paths of the communication graph. In the statement of the theorem, \emph{almost any} means all choices of entries in $A$ except a set of Lebesgue measure zero. These bad values are corresponding to the solutions of a polynomial function \cite{DioComWou03}.

\subsection{Partial Reconstruction}

Reconstructing the subgradients of all other nodes is severely constrained by the topology of the communication graph, the malicious node may turn to reconstruct the subgradients from \emph{some} of the nodes. A logical step forward from the full reconstruction problem is \emph{partial} reconstruction. That is, given a set of nodes, what are the topological requirements for the communication graph that allows a malicious node to reconstruct the subgradients of this set of nodes.

Suppose a malicious node wants to reconstruct the subgradients of a set of nodes $\Ncal$. For the purpose of analysis, we break the input $\Gtilde_t$ of the system $\Scal$ into two parts. One part $\Gtilde_t^\Ncal$ is the columns of $\Gtilde_t$ that are corresponding to the subgradients of the nodes in $\Ncal$, and another part $\Gtilde_t^\Ucal$ is corresponding to all other nodes. The dynamics of the algorithm can be described by the following system $\Scal'$, which is equivalent to the system $\Scal$.
\begin{align}
\label{eq:LTI-representation-partial}
\Scal':
\begin{cases}
W_{t+1} = W_{t} A + \Gtilde_{t}^\Ncal B_\Ncal + \Gtilde_t^\Ucal B_\Ucal \\
Y_{t} = W_{t} C
\end{cases}
\end{align}
$B_\Ncal$ and $B_\Ucal$ are suitable matrices that align the input to corresponding columns. Instead of considering the invertibility of $\Scal'$, we consider the partial invertibility of $\Scal'$---inverting only $\Gtilde_{t}^\Ncal$ from the output $Y_t$. The next theorem relates the partial invertibility of $\Scal'$ to the topological properties of the communication graph.

\begin{theorem}
\label{thm:partial-rec}
The necessary and sufficient conditions for the sequence of output vector $Y_t$ at the malicious node $M$ giving rise to a unique sequence of $\Gtilde_t^\Ncal$ for almost any choice of nonzero elements in A are:
\begin{enumerate}[i)]
\item All nodes in $\Ncal$ are connected to $M$.
\item No other nodes are connected to the nodes in $\Ncal$ but not connected to $M$.
\end{enumerate}
\end{theorem}

The proof of sufficiency is a simple corollary of Theorem
\ref{thm:privacy}. If the nodes in $\Ncal$ and $M$ satisfy the
conditions in Theorem \ref{thm:partial-rec}, the nodes of
$\Ncal\cup\{M\}$ form a network that satisfies the full reconstruction
condition in Theorem \ref{thm:privacy}, and $M$ can reconstruct the
subgradients of the nodes in $\Ncal$ by duplicating the linear
combination and local subgradient steps at the node in $\Ncal$. Similar
to the full reconstruction, the only exception for the partial
reconstruction is $\Gtilde_1^\Ncal$, whose recovery depends on the
knowledge of the initial parameters $W_1^\Ncal$. The proof for necessity is significantly harder than that of the full reconstruction, and the long proof is given in the appendix. This theorem confirms our intuition by saying, for a set of nodes $\Ncal$, if they directly provide information to $M$ and there is no other nodes that ``mix" unknown information into this set of nodes, $M$ can reconstruct the subgradients of the nodes in $\Ncal$, otherwise the subgradients can only be determined up to a linear subspace \cite{SunHad09}.

The theory developed above can guide us to examine or design communication networks with privacy-preserving properties. We define a privacy-preserving communication network as the following.
\begin{definition}
\label{def:privacy-network}
We say a communication network $C(A)$ is privacy-preserving \emph{if and only if} the conditions in Theorem \ref{thm:partial-rec} do not hold for any node $M$ and any set of nodes $\Ncal$.
\end{definition}
A set of nodes is called a \emph{vertex cut} of a directed graph $G$ if the removal of these nodes renders the graph \emph{disconnected}. The connectivity $\kappa(G)$ of the graph is the size of the smallest vertex cut. Suppose a communication network is not privacy-preserving, then there exist node $M$ and a set of nodes $\Ncal$ satisfy the conditions in Theorem \ref{thm:partial-rec}. Furthermore, we assume that not all nodes are connected to $M$. Then removing $M$ makes the graph disconnected because there is no path from the nodes in $\Ucal$ to the nodes in $\Ncal$, so $\{M\}$ is a vertex cut and $\kappa(C(A)) = 1$. The above analysis can be summarized by the following theorem.
\begin{theorem}
\label{thm:connectivity}
For a communication network $C(A)$, if $\kappa(C(A)) > 1$ and $\forall$ node $j$, $|N(j)|<m-1$, then $C(A)$ is privacy-preserving.
\end{theorem}
It can be shown that many interesting networks, including those studied by \citet{DucAgaWai10}, are privacy-preserving. For example, (a) the grid, where nodes are aligned on a 2-dimension grid and connected to the nearest 4 neighbors; (b) the k-dimension hyper-cube, where nodes are placed on the vertices of an imaginary k-dimension hyper-cube, and connected to the neighbor vertices; (c) expander graphs, one can construct expander graphs to have large connectivity. These graphs have good mixing properties.

\subsection{Reconstruction under Projection}

We define auxiliary variables $r_t^i = w_t^i - \what_t^i$ and define $R_t=[r_t^1,\ldots,r_t^m]$. Suppose again that the malicious node is interested in the node in the set $\Ncal$, the dynamics of the distributed online learning algorithm \emph{with projection} can be described by the following system $\Scal''$
\begin{align}
\label{eq:LTI-representation-projection}
\Scal'':
\begin{cases}
W_{t+1} = W_{t} A + \Gtilde_{t}^\Ncal B_\Ncal + [R_{t+1}^\Ncal, \Gtilde_t^\Ucal, R_{t+1}^\Ucal] \begin{bmatrix}B_\Ncal \\ B_\Ucal \\ B_\Ucal\end{bmatrix}\\
Y_{t} = W_{t} C
\end{cases}
\end{align}
Note that reconstructing $\Gtilde_t^\Ucal + R_{t+1}^\Ucal$ in system $\Scal''$ is the same as reconstructing $\Gtilde_t^\Ucal$ in system $\Scal'$, and it has been addressed in Theorem \ref{thm:partial-rec}. Therefore, in order to reconstruct the subgradients $\Gtilde_t^\Ucal$ in system $\Scal''$, it is sufficient to reconstruct or separate the projection difference $R_{t+1}^\Ncal$ from $\Gtilde_{t}^\Ncal$.

Under the formulation of $\Scal''$, we consider $\eta_t g_t^i$ and $r_{t+1}^i$ as \emph{two} separate inputs to the node (learner) $i$, but each node simply propagates the summation $-\eta_t g_t^i+r_{t+1}^i$. For certain types of convex sets, such as hyper-balls or polytopes, it is easy to find different data vectors having the same projection value. It is hard to separate $\Gtilde_t^\Ncal$ and $R_{t+1}^\Ncal$. Formally, we have the following theorem and the proof can be found in the appendix.
\begin{theorem}
\label{thm:projection-rec}
In system $\Scal''$, the output sequence $Y_t$ cannot determine a unique sequence of subgradients $\Gtilde_t^\Ncal$ for any communication network. 
\end{theorem}
The proof of the above theorem follows a similar line of that of Theorem \ref{thm:partial-rec} except different topological arguments. Theorem \ref{thm:projection-rec} should be exercised with caution. It is possible to gain information about the subgradients in the presence of \textit{a priori} knowledge. For example, if $\Omega$ is a $\ell_2$ ball, $\eta_t g_t^i$ and $r_{t+1}^i$ are co-linear, so the summation $-\eta_t g_t^i+r_{t+1}^i$ can determine $g_t^i$ up to a constant factor.

\begin{figure}[thb]
\center{
\subfigure[]{\includegraphics[scale=0.28]{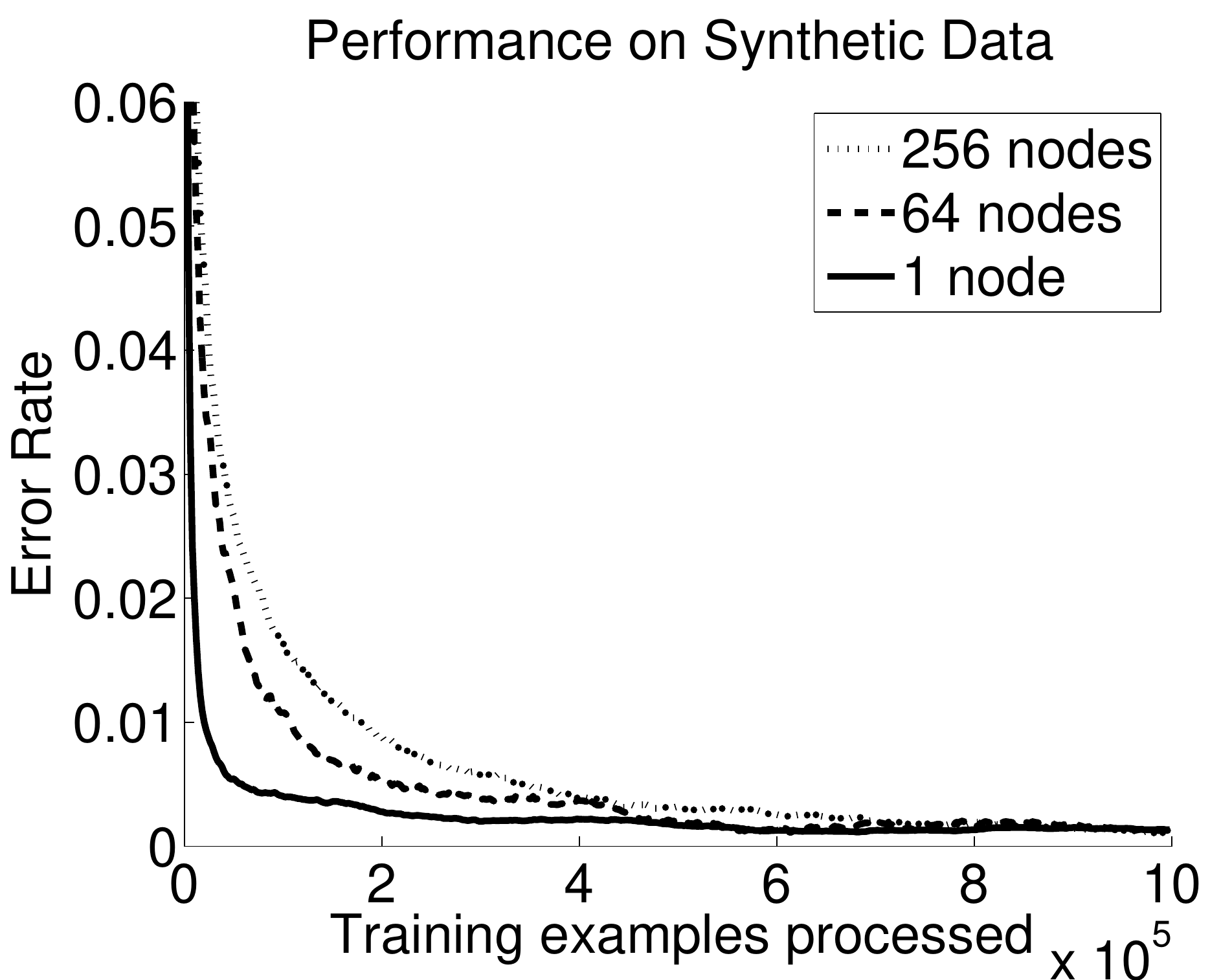}}
\subfigure[]{\includegraphics[scale=0.28]{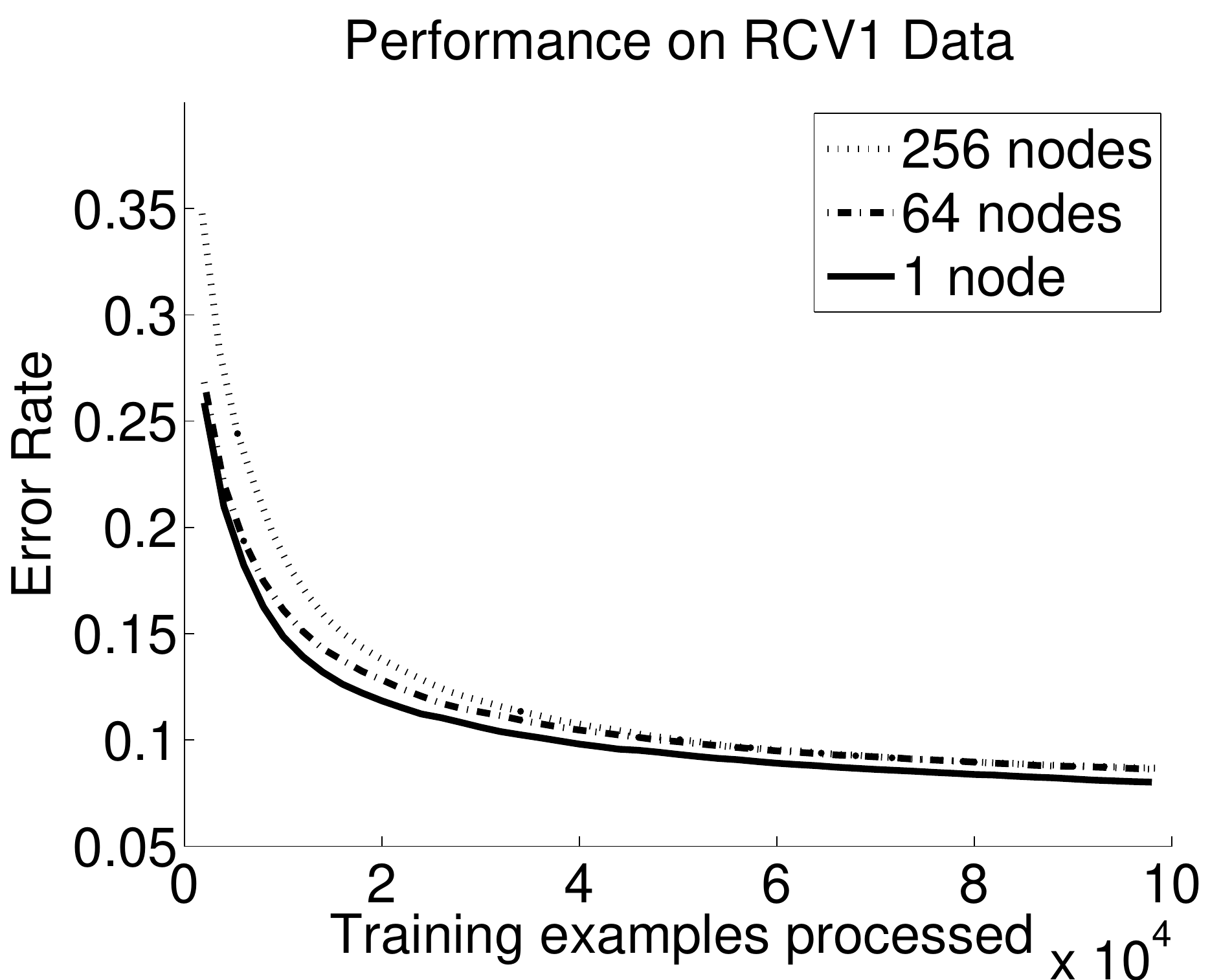}}\\
\subfigure[]{\includegraphics[scale=0.28]{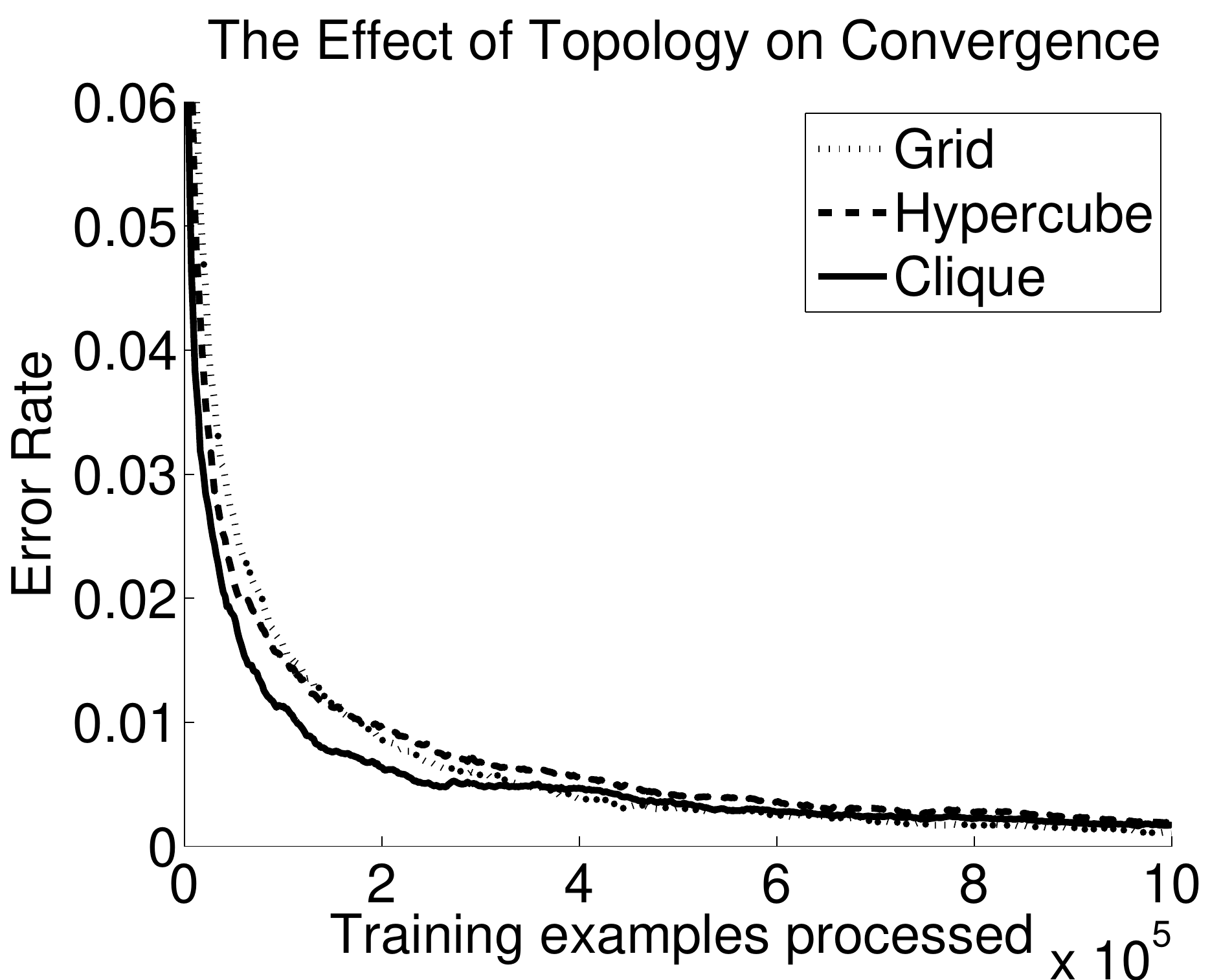}}
}
\caption{ (a) and (b): Convergence of distributed learning on synthetic and real datasets. On both datasets, our distributed online learning algorithm uses up to 256 nodes linked by hypercubes. It converges to the test error rate of sequential online learning. (c)  Convergence of distributed learning with different communication graphs consisting of 256 nodes on synthetic data.  When the communication graphs are \textit{grids} or \textit{hypercubes}, the algorithm converges slightly slower than when the communication graphs are \textit{cliques}. But unlike cliques, grids and cliques prevent malicious nodes from reconstructing subgradients of other nodes.}
\label{fig:simulation}
\end{figure}

\section{Related Works}


Recently some research effort has been devoted to devising  distributed online learning.
For instance \citet{LanSmoZin09} shows that one can distribute the data on slave nodes. The slaves periodically poll the \emph{centralized} master node to
receive the latest parameter vector. This is used to compute stochastic
gradients which are then fed back to the master node at the expense of using delayed subgradients.  Their bounds have the form $O(\sqrt{\tau N})$ and $O(\tau + \tau \log(N/m))$, where $\tau$ is the delay in the subgradient calculation. Given the fact that $\tau$ is as large as $m$ in a round-robin fashion communication scheme, the bounds of \citet{LanSmoZin09} are similar to ours.




The \emph{decentralized} learning paradigm was pioneered in distributed optimization. For example, \citet{DucAgaWai10} proposed a dual averaging algorithm for distributed convex optimization. They provided sharp bounds on their convergence rates as a function
of the network size and topology by careful mixing time arguments. \citet{ZinWeiSmoLi10} proposed to perform local stochastic gradient descent individually then give the output as the average of local parameters at the final step. However, their fixed step size assumption does not guarantee the algorithm to converge to the true optimum. In terms of algorithmic structures and underlying mathematical foundations, our algorithm is a natural extension of the works of \citet{NedOzd09} and \citet{RamNedVee09} for distributed convex optimization to online learning, but our analysis handles \emph{strongly} convex function and yields $O(\log{T})$ regret. If our regret bounds are converted to convergence rates, then we obtain not only $O(1/\epsilon^2)$ rates for convex functions, but also $O(1/\epsilon)$ rates for strongly convex functions, which are not covered by \emcite{NedOzd09,RamNedVee09}. Except the work of \citet{ZinWeiSmoLi10}, which is obviously privacy-preserving due to the lack of communication, \emph{none} of these works considered the privacy-preserving aspect of the algorithms.

Privacy-preserving has been an active research area in machine learning and data mining. Most privacy-preserving machine learning algorithms modify the original algorithms with cryptographic tools to achieve privacy preservation. Two popular techniques are secure multi-party computation (SMC) and randomization. For example, the privacy-preserving versions of linear regression \cite{VaiCliZhu05}, belief propagation/Gibbs sampling \cite{KeaTanWor07} and online prediction over discrete values \cite{SakAra10} use SMC to securely compute function values over distributed data without disclosing them to unwanted identities; the privacy-preserving logistic regression \cite{ChaMon09} uses randomized perturbation to modify the cost function to preserve data privacy. Many algorithms, such as association rule mining and decision tree, can use either SMC or randomization to achieve privacy preservation \cite{VaiCliZhu05}. Compared to the algorithms using SMC and randomization, our analysis on privacy does \emph{not} require any modification of the original algorithm. The privacy-preserving properties of ours are \emph{intrinsic} in the sense that it only relies on a component of our algorithm, the communication graph, to prevent disclosure of local subgradients (hence data) to other nodes.

By treating local parameter $w_t^i$ as an aggregated vector of local subgradients (data), our approach to privacy preservation is closely related to the aggregation-based methods on a conceptual level. For example, \citet{Rup10} trains support vector machines by using group probability over subsets of data. \citet{AviBut07} proposed a boosting based privacy-preserving face detection algorithm by restricting the learner to use limited features provided by the data feeder. One drawback of these algorithms is they sacrifice algorithm performance for data privacy by only revealing aggregated or limited information. By contrast, our algorithm achieves the same asymptotic convergence rate as the sequential algorithm on a fixed number of learners.


\section{Simulations}

We conduct two set of simulations to illustrate how quickly the generalization error of our distributed learning algorithm converges given certain number of nodes and to examine the impact of the topology of communication graphs on the convergence rate. For our implementations, each $f_t^i(w)$ has the form $h(y_t^i\inner{w}{x_t^i})$, where $\{(x_t^i,y_t^i) \in \RR^n \times \{\pm 1\}\}$ are the training data available only to the $i^{th}$ node, and $h(\chi)$ is the hinge loss function $h(\chi) = \max\{1-\chi,0\}$. For robustness, we set the learning rate $\eta_t = \frac{1}{2\sqrt{t}}$.

First, we investigate how the number of nodes affects the predictive performance of our algorithm on both synthetic and RCV1 datasets\footnote{\url{http://www.csie.ntu.edu.tw/~cjlin/libsvmtools/datasets/binary.}}. The synthetic data are generated uniformly from a 10-dimension unit ball. The classifier is randomly sampled and less than 10\% of the labels based on the true classifier are flipped to the wrong labels. In total, we generate 1,000,000 training and 500,000 test examples. The second dataset is actually a subset of the RCV1 dataset. This subset contains 100,000 training examples, 100,000 test examples, and 47,236 features with many zero entries for each sample.
Figures \ref{fig:simulation}.(a) and (b) summarize the results.  In line with the theoretically guarantee the regret our distributed algorithm converges, the test error of our algorithm, even with 256 nodes, indeed converges to that of the sequential learner on both datasets.

\commenttext{
In the first experiment, we simulate the algorithm for different numbers of nodes and observe the error rate on the test set as more training examples are used to train the weight vectors. The topology of the communication graph is \textit{hypercube}, where two nodes are connected if their indices' Hamming distance is 1 in the binary representation. The combination weights in $A$ are equal for every nodes. Figures \ref{fig:simulation}.(a) and (b) show, when increasing the number of nodes up to 256, the error rates on the test set converge to that of the sequential runs.
}

For the second experiment, we construct three types of communication graphs consisting of 256 nodes: i) \textit{grid} where nodes are laid and connected on a 2-D mesh grid; ii) \textit{hypercube} where nodes are laid and connected on a 8-dimensional hypercube; 3) \textit{clique} where the nodes form a clique. As shown in figure \ref{fig:simulation}(c), the \textit{clique} topology leads to slightly faster convergence than \textit{grid} and \textit{hypercube}, but it discloses subgradients in the presence of malicious nodes according to Theorem \ref{thm:privacy}.

\section{Discussion}
\label{sec:Discussion}


We have only analyzed the case where the communication matrix $A$ is fixed, and does not evolve over time. Our proofs can be extended to the settings of asynchronous update or random communication as studied by \citet{NedOzd09}. The resulting linear systems are time-invariant, which is much harder to analyze. However, we conjecture that all the privacy-preserving properties still hold if the transient network connectivity is greater than one upon any update step.




\appendix
\section{Proofs of the Regret Bounds}

The subgradient (set) $\partial_x f(\cdot)$ of a convex function $f(x)$ at $x_0$ is
defined as 
\begin{align}
  \label{eq:subgradient-def}
  g \in \partial f(x_0) \iff \forall y,~~f(y)-f(x_0) \geq
  \inner{y-x_0}{g}.
\end{align}
A convex function $f(\cdot)$ defined on domain $\Omega$ is said to be
strongly convex with modulus $\lambda>0$ if and only if 
\begin{align}
  \label{eq:strong-convex}
  \forall x, y \in \Omega,~f(y) - f(x)-\inner{y-x}{\partial_x
    f(x)} \geq \frac{\lambda}{2} \nbr{y-x}^2
\end{align}
where $\partial_x f(x)$ is the subgradient. The Euclidean projection operator onto a set $\Omega \subseteq \RR^{n}$ is defined as
\begin{align}
  \label{eq:proj-def}
  P_{\Omega}(w') = \argmin_{w \in \Omega} \nbr{w - w'}.
\end{align}
We define the \emph{average} parameter vector $w_t$ as
\begin{align}
w_t = \frac{1}{m}\sum_{i=1}^m w_t^i
\end{align}
Our proof is based on an analysis of the sequence of values $w_t$.

\subsection{Lemmas}

We start from a key result concerning the decomposition of regret is Lemma \ref{lem:key-online} given below.

\begin{lemma}
  \label{lem:key-online}
  Let $w_{t}^{i}$ denote the sequences generated by Algorithm
  \ref{alg:stoc-grad-desc}. Denote $\bar{g}_{t}^{i} = \partial_{w} f_{t}^{i}(w_{t})$. For
  any $w \in \Omega$ we have
  \begin{align}
    \label{eq:stoc-relation}
    \nbr{w_{t+1}-w}^2 & \leq  (1-2 \eta_t \lambda) \nbr{w_{t}-w}^2 +
    \frac{4\eta_t^2}{m^2} \rbr{\sum_{i=1}^m \nbr{g_t^i}}^2 \nonumber\\
    &- \frac{2
      \eta_t}{m} (f_t(w_t) - f_t(w)) \nonumber \\
    & + \frac{2\eta_{t}}{m} \sum_{i=1}^{m} (\nbr{g_{t}^{i}} +
    \nbr{\bar{g}_{t}^{i}}) \nbr{w_{t} - w_{t}^{i}} \nonumber\\
    &+ \frac{2\eta_{t}}{m}
    \sum_{i=1}^{m} \nbr{g_{t}^{i}} \nbr{w_{t} - \hat{w}_{t+1}^{i}}
  \end{align}
\end{lemma}

\begin{proof}
Define
\begin{align}
  \label{eq:rdef}
  r_{t}^{i} := w_{t}^{i} - \hat{w}_{t}^i =
  P_{\Omega}\rbr{\hat{w}_{t}^{i}} - \hat{w}_{t}^{i}.
\end{align}
Recall that $\Omega$ is assumed to be convex, $A$ is a doubly stochastic
matrix, and $w_{t}^{j} \in \Omega$ for all $j$.  Therefore, $A_{ji}
\geq 0$, $\sum_{j} A_{ji} = 1$, and $\sum_{j} A_{ji} w_{t}^{j} \in
\Omega$ for all $i$.  By this observation, the definition of the
projection operator \eqref{eq:proj-def}, and the definition of
$\hat{w}_{t+1}^i$ in Line 6 of Algorithm \ref{alg:stoc-grad-desc} we
have the following estimate for the norm of $r_{t+1}^{i}$
\begin{align}
  \label{eq:r-normbound}
  \nbr{r_{t+1}^i} &= \nbr{P_{\Omega}\rbr{\hat{w}_{t+1}^i} -
    \hat{w}_{t+1}^i} \nonumber \\
    &\leq \nbr{\sum_{j}A_{ji} w_{t}^j - \hat{w}_{t+1}^i}
  = \eta_{t}\nbr{g_{t}^i}
\end{align}

Then, we define the following matrices to simplify the notations.
  \begin{align*}
    & W_t=[{w_t^1},\ldots,{w_t^m}],~~~~\hat{W}_t=[{\hat{w}_t^1},\ldots,{\hat{w}_t^m}] \\
    & G_t=[g_t^1,\ldots,g_t^m],~~~~R_t=[r_t^1,\ldots,r_t^m]
  \end{align*}
  Since $A$ is doubly stochastic $Ae = 1$. Therefore, by using
  \eqref{eq:rdef} and the update in step 6 of
  Algorithm~\ref{alg:stoc-grad-desc}, we have the relation
  \begin{align}
    \label{eq:wtplusone}
    {w_{t+1}} & = \frac{1}{m} W_{t+1} e = \frac{1}{m} (W_{t} A -
    \eta_{t} G_{t} + R_{t+1}) e \nonumber\\
    \nonumber & = \frac{1}{m} W_{t} e - \frac{\eta_{t}}{m} G_{t} e +
    \frac{1}{m} R_{t+1} e \\
    &= w_{t} - \frac{\eta_{t}}{m} \sum_{i=1}^{m}
    g_{t}^{i} + \frac{1}{m} \sum_{i=1} r_{t+1}^{i}.
  \end{align}
  Using the above relation we unroll $\nbr{w_{t+1}-w}^2$ by
  \begin{align}
    \label{eq:stoc-relation-naive}
    \nbr{w_{t+1}-w}^2 & = \nbr{w_t-w}^2 + \frac{1}{m^{2}}
    \nbr{\sum_{i=1}^m \rbr{r_{t+1}^i+\eta_t g_t^i}}^2 \nonumber \\
    & - \frac{2 \eta_t}{m} \sum_{i=1}^m \inner{g_t^i}{w_t-w} +
    \frac{2}{m} \sum_{i=1}^m \inner{r_{t+1}^i}{w_t-w}.
  \end{align}
  In view of \eqref{eq:r-normbound}
  \begin{align}
    \label{eq:stoc-relation-term1}
    \frac{1}{m^2} \nbr{\sum_{i=1}^m \rbr{r_{t+1}^i+\eta_t g_t^i}}^2 & \leq
    \frac{1}{m^2} \rbr{\sum_{i=1}^m \nbr{r_{t+1}^i} + \eta_t \nbr{g_t^i}}^2 \nonumber\\
    &=\frac{4\eta_t^2}{m^2} \rbr{\sum_{i=1}^m \nbr{g_t^i}}^2.
  \end{align}
  Next we turn our attention to the $$-\sum_{i} \inner{g_{t}^{i}}{w_{t}-
    w}$$ term which we bound using \eqref{eq:subgradient-def} and \eqref{eq:strong-convex} as follows:
  \begin{align*}
    &-\inner{g_{t}^{i}}{w_{t} - w} = -\inner{g_{t}^{i}}{w_{t} -
      w_{t}^{i}}
    - \inner{g_{t}^{i}}{w_{t}^{i} - w} \\
    & \leq \nbr{g_{t}^{i}}\nbr{w_{t} - w_{t}^{i}} + f_{t}^{i}(w) -
    f_{t}^{i}(w_{t}^{i})  - \lambda \nbr{w_{t}^i - w}\\
    & = \nbr{g_{t}^{i}}\nbr{w_{t} - w_{t}^{i}} + f_{t}^{i}(w_{t}) -
    f_{t}^{i}(w_{t}^{i}) - \lambda \nbr{w_{t}^i - w}\\
    &~~~+f_{t}^{i}(w) - f_{t}^{i}(w_{t})  \\
    & \leq \nbr{g_{t}^{i}}\nbr{w_{t} - w_{t}^{i}} +
    \inner{\bar{g}_{t}^{i}}{w_{t}- w_{t}^i} - \lambda \nbr{w_{t}^i -
      w_{t}}\\
    &~~~- \lambda \nbr{w_{t}^i - w} +f_{t}^{i}(w) - f_{t}^{i}(w_{t}) \\
    & \leq \rbr{\nbr{g_{t}^{i}} + \nbr{\bar{g}_{t}^{i}}} \nbr{w_{t} -
      w_{t}^{i}} \\
    &~~~- \lambda \nbr{w_{t} - w} + f_{t}^{i}(w) -
    f_{t}^{i}(w_{t}) .
\end{align*}
The last inequality is by using $$\inner{\bar{g}_{t}^{i}}{w_{t}-
  w_{t}^{i}} \leq \nbr{\bar{g}_{t}^{i}} \nbr{{w_{t}- w_{t}^{i}}}$$ and
$$\nbr{w_{t}^{i} - w_{t}} + \nbr{w_{t}^{i} - w} \geq \nbr{w_{t} - w}$$
Summing up over $i=1,\ldots,m$, obtains
\begin{align}
  \label{eq:stoc-relation-term2}
  -\sum_{i=1}^m \inner{g_{t}^{i}}{w_{t} - w}  \leq &\sum_{i=1}^m
  \rbr{\nbr{g_{t}^{i}} + \nbr{\bar{g}_{t}^{i}}} \nbr{w_{t} - w_{t}^{i}} \nonumber\\
  &- \lambda m \nbr{w_{t} - w} - \rbr{f_{t}(w_{t}) - f_{t}(w)}
\end{align}

The projection operator satisfies the following property
\begin{align}
  \label{eq:proj-non-expansive}
  \inner{P_\Omega \rbr{\hat{w}}-\hat{w}}{\hat{w}-w} \leq -\nbr{P_\Omega
    \rbr{\hat{w}}-\hat{w}}^2 \leq 0,\;\; \forall w \in \Omega.
\end{align}

In order to estimate $\inner{r_{t+1}^i}{w_t-w}$, we use \eqref{eq:rdef},
\eqref{eq:proj-non-expansive}, and \eqref{eq:r-normbound} to write
\begin{align}
  \label{eq:stoc-relation-term3}
  \inner{r_{t+1}^i}{w_t-w} & = \inner{r_{t+1}^i}{w_t-\hat{w}_{t+1}^i} \nonumber\\
  &~~~+ \inner{P_{\Omega}\rbr{\hat{w}_{t+1}^i} - \hat{w}_{t+1}^i}{\hat{w}_{t+1}^i-w} \nonumber \\
  & \leq \inner{r_{t+1}^i}{w_t-\hat{w}_{t+1}^i} \nonumber\\
  & \leq \eta_t
  \nbr{g_{t}^i}\nbr{w_t-\hat{w}_{t+1}^i}.
\end{align}
Combining \eqref{eq:stoc-relation-term1}, \eqref{eq:stoc-relation-term2}
and \eqref{eq:stoc-relation-term3} with \eqref{eq:stoc-relation-naive}
completes the proof.
\end{proof}

The projection operator satisfies the following property
\begin{align}
  \label{eq:proj-non-expansive}
  \inner{P_\Omega \rbr{\hat{w}}-\hat{w}}{\hat{w}-w} \leq -\nbr{P_\Omega
    \rbr{\hat{w}}-\hat{w}}^2 \leq 0,\;\; \forall w \in \Omega.
\end{align}

The following lemma to upper bound the terms $\nbr{{w_t}-{w_t^i}}$ and $\nbr{{w_t}-{\hat{w}_{t+1}^i}}$ in \eqref{eq:stoc-relation}. The convergence rate in \eqref{eq:markov-def} plays a central role in this lemma.

\begin{lemma}
  \label{lem:bnd}
  If the assumptions in section
  \ref{sec:Coopautononline} hold, and let $\beta$ be as in \eqref{eq:markov-def}, then
  \begin{align}
    \label{eq:difference1}
    \nbr{w_{t}-w_{t}^{i}} \leq 4 L \sum_{k=1}^{t-1} \eta_{t-k} \beta^{k-1} \\
    \label{eq:difference2}
    \nbr{w_{t}-\hat{w}_{t+1}^{i}} \leq 4 L \sum_{k=0}^{t-1} \eta_{t-k} \beta^{k}.
  \end{align}
\end{lemma}

\begin{proof}

Using the notations defined in the proof of Lemma~\ref{lem:key-online}, we unroll the relation
\begin{align}
W_{t}=W_{t-1} A - \eta_{t} G_{t-1} + R_{t}
\end{align}
which is defined through Algorithm \ref{alg:stoc-grad-desc} yields
  \begin{align}
    \label{eq:recursive-relation}
    W_t = W_1 A^{t-1} - \sum_{k=1}^{t-1} \eta_{t-k} G_{t-k} A^{k-1} +
    \sum_{k=1}^{t-1} R_{t-k+1} A^{k-1}.
  \end{align}
  Using $A^{k} e = 1$ for all $k$, \eqref{eq:markov-def},
  \eqref{eq:r-normbound}, and the above relation we can write

  \begin{align*}
    &\nbr{w_t-w_t^i} = \nbr{W_t\rbr{\frac{1}{m} e-e_i}} \\
    & \leq \nbr{w_1 - w_1^i} + \sum_{k=1}^{t-1} \eta_{t-k}
    \nbr{G_{t-k}\rbr{\frac{1}{m}e-A_i^{k-1}}} \\
    &~~~ + \sum_{k=1}^{t-1} \nbr{R_{t-k+1} \rbr{\frac{1}{m}e-A^{k-1}_i}} \\
    & \leq 4 L \sum_{k=1}^{t-1} \eta_{t-k} \beta^{k-1}
  \end{align*}
  We omit the proof for \eqref{eq:difference2} which follows along
  similar lines.

\end{proof}

A general lemma on the regret bounds is the following

\begin{lemma}
  \label{lem:regret-online}
  Let $w^{*} \in \Omega^{*}$ denote the best parameter chosen in
  hindsight. Then the regret of Algorithm~\ref{alg:stoc-grad-desc} can
  be bounded via
  \begin{align}
    \label{eq:regret}
    \sum_{t=1}^{T} f_{t}(w_{t}^j) - f_{t}(w^{*}) & \leq
    mF\rbr{\frac{1}{2 \eta_{T}} - T \lambda} \nonumber \\ & + 4mCL^{2}\sum_{t=1}^{T} \eta_{t},
  \end{align}
where $C$ is a communication-graph-dependent constant defined as
\begin{align}
  \label{eq:markov-constant}
  C=\frac{5-\beta}{1-\beta}.
\end{align}
\end{lemma}

\begin{proof}
Set $w=w^{*}$, divide both sides of
  \eqref{eq:stoc-relation} by $\frac{2 \eta_t}{m}$ and rearrange to
  obtain
  \begin{align*}
    & f_t(w_t) - f_t(w^{*}) \\
    = & f_t(w_t^j) - f_t(w^{*}) + f_t(w_t) - f_t(w_t^j) \\
    \leq \; & \; \frac{m}{2
      \eta_t}\sbr{(1-2\eta_t\lambda)\nbr{w_t-w^{*}-\nbr{w_{t+1}-w^{*}}}} \\
      & + 2 \frac{\eta_t}{m} \rbr{\sum_{i=1}^{m} \nbr{g_t^i}}^2 + 2L\sum_{i=1}^m \nbr{w_t-w_t^i} \\
      & + L \sum_{i=1}^m \nbr{w_t-\hat{w}_{t+1}^i} + mL \nbr{w_t-w_t^j}
  \end{align*}

  Plug in the estimate of the subgradients and the bounds
  \eqref{eq:difference1} and \eqref{eq:difference2}.

  \begin{align*}
    & f_t(w_t) - f_t(w^{*}) \\
    \leq \; &\frac{m}{2
      \eta_t}\sbr{(1-2\eta_t\lambda)\nbr{w_t-w^{*}}-\nbr{w_{t+1}-w^{*}}}
    \\ & + 2 m L^2 \eta_t + 12 L^2 m \sum_{k=1}^{t-1} \eta_{t-k} \beta^{k-1} +
    4 L^2 m \sum_{k=0}^{t-1} \eta_{t-k} \beta^k \\
    \leq \; & \frac{m}{2
      \eta_t}\sbr{(1-2\eta_t\lambda)\nbr{w_t-w^{*}}-\nbr{w_{t+1}-w^{*}}}
    \\
    & + 4 m L^2 \eta_t + 16 L^2 m \sum_{k=1}^{t-1} \eta_{t-k} \beta^{k-1}
  \end{align*}

  Summing over $t=1,\ldots,T$

  \begin{align*}
    & \sum_{t=1}^T f_t(w-t) - f_t(w^*) \\
    \leq \; & m
    \underbrace{\sum_{t=1}^T \frac{1}{2\eta_t}
      \sbr{(1-2\eta_t\lambda)\nbr{w_t-w^{*}}-\nbr{w_{t+1}-w^{*}}}}_{:=
      C_1}
    \\
    & + 4 m L^2 \sum_{t=1}^T \eta_t + 16 L^2 m \underbrace{\sum_{t=1}^T
      \sum_{k=1}^{t-1} \eta_{t-k} \beta^{k-1}}_{:= C_2}
  \end{align*}

  Since the diameter of $\Omega$ is bounded by $F$
  \begin{align*}
    C_1 & = \rbr{\frac{1}{2\eta_1}-\lambda}\nbr{w_1-w^*} -
    \frac{1}{2\eta_T}\nbr{w_{T+1}-w^*} \\
    &~~~ + \sum_{t=2}^{T} \nbr{w_t-w^*}
    \rbr{\frac{1}{2\eta_t}-\frac{1}{2\eta_{t-1}}-\lambda} \\
    & \leq \rbr{\frac{1}{2\eta_1}-\lambda} F + \sum_{t=2}^{T} F
    \rbr{\frac{1}{2\eta_t}-\frac{1}{2\eta_{t-1}}-\lambda} \\
    & = F \rbr{\frac{1}{2\eta_T}-T\lambda}
  \end{align*}
  Let $I(t>k)$ be the indicator function which is $1$ when $t>k$ and $0$
  otherwise. Then
  \begin{align*}
    C_2 \; & = \; \sum_{t=1}^T \sum_{k=1}^{T} \eta_{t-k} \beta^{k-1} I(t>k)
    = \sum_{k=1}^{T} \beta^{k-1} \sum_{t=k+1}^T \eta_{t-k} \\
    & \; \leq \; \sum_{k=1}^{T} \beta^{k-1} \sum_{t=1}^T \eta_{t} \; \leq \;
    \frac{1}{1-\beta} \sum_{t=1}^T \eta_{t}
  \end{align*}
  Plug in the estimate for $C_1$ and $C_2$, to obtain \eqref{eq:regret}.
\end{proof}

\subsection{Proof of Theorem \ref{thm:reg-bound}}

First consider $\lambda > 0$ with $\eta_{t} = \frac{1}{2 \lambda t}$. In this case $\frac{1}{2 \eta_{T}} = T\lambda$, and consequently \eqref{eq:regret} in Lemma~\ref{lem:regret-online} specializes to
\begin{align*}
\sum_{t=1}^{T} f_{t}(w_{t}) - f_{t}(w^{*}) & \leq \frac{C L^{2} m}{2\lambda} \sum_{t=1}^{T} \frac{1}{t} \\
&\leq \frac{C L^{2} m}{2\lambda} (1 + \log(T)).
\end{align*}

When $\lambda = 0$, and we set $\eta_{t} = \frac{1}{2 \sqrt{t}}$ and to rewrite \eqref{eq:regret} as
\begin{align*}
\sum_{t=1}^{T} f_{t}(w_{t}) - f_{t}(w^{*}) & \leq m F \sqrt{T} +
C L^{2}m \sum_{t=1}^{T} \frac{1}{2\sqrt{t}} \\
& \leq m F \sqrt{T} + C L^{2} m \sqrt{T}.
\end{align*}

\section{Generalization Bound}

We investigate the relationship between the regret bounds and the generalization ability of the proposed algorithms. Let $\Fcal$ be the space of all possible choices of $f_t^i(w)$ equipped with a probability measure. Random variables are denoted as capital letters, \eg~$f_t^i(w)$ is a realization of the random variable $F_t^i(w) \in \Fcal$. We assume the functions $f_t^i(w)$ are generated as \emph{i.i.d.} random elements in $\Fcal$ according to the unknown distribution over $\Fcal$. The \emph{risk} of $w$ is defined as $\rk(w)=\EE[F(w)]$. A common form of $f_t^i(w)$ in the cost function \eqref{eq:stoc-obj} is $l(y_t^i,\inner{w}{x_t^i})$ where $l(\cdot,\cdot)$ is the loss function. In this case, the risk is the expected loss when the parameter is $w$. Since the data $x_t^i$ are bounded in most cases, we can assume the loss $l(\cdot,\cdot)$ or the functions $f_t^i$ are bounded. Let $N$ denote the number of all functions $f_t^i$ up to the iteration $T$ and $N=mT$. The following theorem bounds the risk by the regret $\Rcal_{DA}$.

\begin{theorem}
\label{thm:gen-bernstein}
If $\forall~f\in\Fcal$, $|f|<\frac{1}{2}$, then for any $0< \delta \leq 1$, with at least $1-\delta$ probability, we have
\begin{align}
\label{eq:generalization-bound-new}
& \inf_{t=1,\ldots,T} \emph{\rk}(W_t^j) - \min_{w\in\Omega} \emph{\rk}(w)
< \frac{\Rcal_{DA}}{N} \nonumber \\
& + \frac{36}{N}\ln\frac{\Rcal_{DA}+3}{\delta} + \frac{2}{N}\sqrt{\Rcal_{DA}\ln{\frac{\Rcal_{DA}+3}{\delta}}}
\end{align}
\end{theorem}

Note that $W_t^j$ are random since $F_t^j$ are random. The inequality \eqref{eq:generalization-bound-new} gives $O(1/N)$ bound on the risk of the best aggregated parameter for strongly convex functions, which translates to $O(1/\epsilon)$ convergence rate (in probability). The key to the proof of the theorem is the generalization bound for sequential online learning by \citet{CesGen06}, which is based on Bernstein's martingale inequality.

\begin{proof}
Let $\overline{w}_t=(w_t^1,\ldots,w_t^m)$ be the parameter vector at the iteration $t$. Since $w_t$ can be represented by a function of $\overline{w}_t$ by $w_t^j=\frac{1}{m}\overline{w}_t \cdot e_j$, we can define $\overline{f}_t(\overline{w}_t)=\sum_{i=1}^m f_t^i(w_t^j)$. The aggregated risk is defined as
\begin{align}
\label{eq:aggregated-risk}
\overline{\rk}(\overline{w}_t) = \EE[\overline{F}_t(\overline{w}_t)]
\end{align}
Since $F_t^i$ are \emph{i.i.d.}, we have $\overline{\rk}(\overline{w}_t^j)=\sum_{i=1}^m \EE[F_t^i(w_t^j)]=m\cdot\rk(w_t^j)$.

In terms of $\overline{f}_t$ and $\overline{w}_t$, Algorithm \ref{alg:stoc-grad-desc} can be regarded as a sequential online learning algorithm that updates $\overline{w}_t$ with $\overline{f}_t$. This view of the algorithm falls into the general setting of online learning algorithm studied in \citep{CesGen06}, if we further interpret $\overline{w}_t$ as hypotheses and $\overline{f}_t$ as training examples. Proposition 2 of \citep{CesGen06} gives
\begin{align}
&\PP\left(\frac{1}{T}\sum_{t=1}^{T} \overline{\rk}(\overline{W}_t\right) < \frac{\Rcal_{DA}}{T} + \frac{1}{T}\EE\left[\min_{w\in\Omega}\sum_{t=1}^T \sum_{i=1}^m F_t^i(w)\right]\nonumber \\
&\frac{36}{T}\ln\frac{\Rcal_{DA}+3}{\delta} + 2\sqrt{\frac{\Rcal_{DA}}{T^2}\ln{\frac{\Rcal_{DA}+3}{\delta}}})>1-\delta
\end{align}
The theorem follows by recognizing the fact $\overline{\rk}(\overline{W}_t)=m \cdot \rk(W_t^j)$ and
\begin{align*}
& \EE\left[\min_{w\in\Omega}\sum_{t=1}^T \sum_{i=1}^m F_t^i(w)\right] <
\min_{w\in\Omega}\EE\left[\sum_{t=1}^T F_t(w)\right] \\
& = \min_{w\in\Omega} T \cdot \EE\left[F(w)\right] = T \min_{w\in\Omega} \rk(w).
\end{align*}
\end{proof}

\section{Proofs of the Privacy-Preserving Results}

\subsection{Proof of Theorem \ref{thm:privacy}}

A \emph{path} $p$ from node $i_0$ to node $i_\tau$ is a sequence of nodes $i_0,i_1,\ldots,i_\tau$, and $(i_j,i_{j+1})$ is an edge for every $0 \geq j <\tau$. Two paths $p_1$ and $p_2$ are \emph{disjoint} if they have no common nodes. A set of paths are disjoint if they are pairwise disjoint. Let $\Xcal_1$ and $\Xcal_2$ are two sets of nodes, a $r-linking$ between $\Xcal_1$ and $\Xcal_2$ are a set of $r$ disjoint paths that start in $\Xcal_1$ and end in $\Xcal_2$ \citep{SunHad09}.

We apply Theorem 1 in \citep{SunHad09} to the system described by $\Scal$, where $M$ can reconstruct the input, \ie~gradients $G_t$, if and only if there exists a $m-linking$ from all nodes $\{1,2,\ldots,m\}$ to $M$ and its neighbors. However, this is only possible when every nodes are neighbors of $M$ and the paths of the $m-linking$ are the nodes themselves.

\subsection{Proof of Theorem \ref{thm:partial-rec}}

As sufficiency is straight-forward, we prove the necessity here. For a sequence of real vectors $\{A_t\}_{t=1}^\infty$, the \emph{one-side $z$-transform} is defined as
\begin{align}
A(z)=\sum_{t=0}^\infty z^{-t} A_{t+1},~~~z \in \CC
\end{align}
where $A(z)$ is well-defined in the complex plane except of a disk centered at zero. The relation between the $z$-transforms of the variables in the system $\Scal$ (equivalently $\Scal'$) is
\begin{align}
Y(z)=W_1 (zI-A)^{-1}Cz + \Gtilde(z) \underbrace{(zI-A)^{-1}C}_\textrm{transfer matrix
}
\end{align}
where the transfer matrix of the system $\Scal'$ is defined as
\begin{align}
T(z)=(zI-A)^{-1}C=\begin{bmatrix}B_\Ncal(zI-A)^{-1}C \\ B_\Ucal(zI-A)^{-1}C\end{bmatrix}=\begin{bmatrix}T_\Ncal(z) \\ T_\Ucal(z)\end{bmatrix}
\end{align}
Each element of $T(z)$ is a rational function and the matrix rank is taken over the rational expression field. We further assume that $W_1=0$, otherwise it may be absorbed into the first input $\Gtilde_1$. The readers may find more detailed description of the above definitions and concepts in standard textbooks on modern control theory, \eg~\citep{Bro91}. The proof is divided into two steps.

\textbf{Step 1:} Supposing $|\Ncal|=r$, we first show that if $Y_t$ determines a unique sequence of inputs to the nodes in $\Ncal$, we must have
\begin{align}
\label{eq:rank-condition}
\rank(\begin{bmatrix}T_\Ncal(z) \\ T_\Ucal(z)\end{bmatrix})-\rank(T_\Ucal(z))=r
\end{align}
We prove this by contradiction. Suppose \eqref{eq:rank-condition} does not hold. Then there exists at least one row of $T_\Ncal(z)$ that is linearly dependent on the other rows of $T(z)$.  Let $T_\Ncal^i(z)$ be this linearly dependent row. Then, there exists a vector $\Gtilde(z)$, with the $i$-th element nonzero such that $\tilde{G}(z)T(z) = 0$. This corresponds to a nonzero input at one of the nodes in $\Ncal$, but the output $Y_t$ is zero for all time, and thus this nonzero input cannot be recovered.

\commenttext{
Then there exists a submatrix $T'(z)$ of the transfer matrix $T(z)$ consisting of $r$ rows of $T(z)$, such that $T'(z)$ has at least one row from $T_\Ucal(z)$ and $\rank(T'(z))=r$. We denote the set of nodes associated with the rows of $T'(z)$ as $\Mcal$.

The choice of $T'(z)$ ensures that there exists a row of $T_\Ncal(z)$ that doesn't in $T'(z)$, let us denote the row as $T^i_\Ncal(z)$, and the corresponding element in $\Gtilde_t$ as $\Gtilde_t^i$. We assign an arbitrary non-zero fixed sequence of values to $\Gtilde_t^i$ and assign sequences of zero to inputs that are not in $\Mcal$. The relation of the variables in the $z$-transform domain reads
\begin{align}
\label{eq:z-transform-zero}
Y(z)+\Yhat(z)&= \Gtilde^\Mcal(z) \; T'(z) \\
\Yhat(z) &= -\Gtilde_t^i\;T^i(z)
\end{align}
Let the sequence of values generating $\Yhat(z)$ through $z$-transform be $\{\Yhat_t\}$. By the results on system inversion \citep{Wil74,SunHad09}, there exists a sequence of $\Gtilde_t^\Mcal$ such that \eqref{eq:z-transform-zero} holds when $Y(z)=0$. Therefore we construct a sequence of input $\Gtilde_t$ with non-zero $\Gtilde_t^\Ncal$ that can give rise to zero output. This sequence of inputs is indistinguishable from zero input, thus the malicious node $M$ cannot identify a unique sequence of inputs to the system.
}

\textbf{Step 2:} We relate the rank condition \eqref{eq:rank-condition} to the topology of the communication graph in this step and complete the proof.

Let us denote the set of the neighbor nodes of $M$ as $\Pcal$. According to \citep{SunHad09} and \citep{DioComWou03}, the rank of the transfer matrix of $\Scal'$ can be analyzed under the framework of \emph{structured} systems. Given a graph, for any choice of nonzero elements in A except for a set of measure zero,
\begin{align*}
\rank(T(z)) = \; &\textrm{max. \# of vertex disjoint paths} \\
& \textrm{from all nodes to }\{M\}\cup\Pcal \\
\rank(T_\Ucal(z)) = \; &\textrm{max. \# of vertex disjoint paths} \\
& \textrm{from }\Ucal\textrm{ to }\{M\}\cup\Pcal
\end{align*}
It is obvious that $\rank(T(z))=\deg(M)+1$ where $\deg(M)$ is the degree of $M$, as we may choose the vertex disjoint paths to be the nodes in $\{M\}\cup\Pcal$ themselves. We denote $\rank(T_\Ucal(z))=u$. The rank condition \eqref{eq:rank-condition} reads
\begin{align}
\label{eq:topology-condition}
\deg(M)+1=r+u
\end{align}
First, partition the set $\{M\}\cup\Pcal$ as $\{M\}\cup\Pcal-\Ncal$ and $\{\{M\}\cup\Pcal\}\cap{N}$. Thus
\begin{align}
& \deg{M}+1 = \{M\}\cup\Pcal \nonumber \\
& = |\{M\}\cup\Pcal-\Ncal| + |\{\{M\}\cup\Pcal\}\cap{N}|.
\end{align}
Now, if N is not contained in $\{M\}\cup\Pcal$, then we have $ |\{\{M\}\cup\Pcal\}\cap{N}| < |\Ncal| = r$.  Furthermore, since $\{M\}\cup\Pcal-\Ncal$ is a subset of $\Ucal$, we have $u \ge |\{M\}\cup\Pcal-\Ncal|$.  Thus we would have $\deg{M}+1 < u + r$, which contradicts \eqref{eq:topology-condition}. Thus we must have $\Ncal$ being a subset of $\{M\}\cup\Pcal$.

Next, suppose that some node in $\Ncal$ has a neighbor in $\Ucal$ that is not also in $\{M\}\cup\Pcal$.  Then we have $u > |\{M\}\cup\Pcal-\Ncal|$, and since $|\Ncal| = r$ (which means that $|\{\{M\}\cup\Pcal\}\cap{N}| = |\Ncal| = r$), we have $\deg{M}+1 < u + r$, which again contradicts \eqref{eq:topology-condition}. Thus, no node in $\Ncal$ can have a neighbor that is not in $\{M\}\cup\Pcal$.

%
%


\subsection{Proof of Theorem \ref{thm:projection-rec}}

The inputs are $\Gtilde_t$ and $R_t$. Let $B_{\Ucal'}=[B_\Ucal^T,B_\Ncal^T,B_\Ucal^T]^T$ the transfer matrix of $\Scal''$ is
\begin{align*}
T(z)=\begin{bmatrix}(zI-A)^{-1}C \\ (zI-A)^{-1}C \end{bmatrix}=\begin{bmatrix}B_\Ncal(zI-A)^{-1}C \\ B_{\Ucal'}(zI-A)^{-1}C\end{bmatrix}
=\begin{bmatrix}T_\Ncal(z) \\ T_{\Ucal'}(z)\end{bmatrix}
\end{align*}
Similar to step 1 in the proof of Theorem \ref{thm:partial-rec}, the output sequence $Y_t$ determines a unique sequence of subgradient inputs $\Gtilde_t$ to the nodes in $\Ncal$ \emph{if and only if}
\begin{align}
\label{eq:rank-condition-proj}
\rank(\begin{bmatrix}T_\Ncal(z) \\ T_{\Ucal'}(z)\end{bmatrix})-\rank(T_{\Ucal'}(z))=r
\end{align}
Next, we relate the rank condition \eqref{eq:rank-condition-proj} to the topological property of the communication graph. We construct a directed graph $C'(A)$ by adding two input nodes $i_g$ and $i_r$ for each node (learner) $i$ in the communication graph $C(A)$ and two edges $(i_g, i)$ and $(i_r, i)$. The two input nodes are corresponding to $\eta_t g_t^i$ and $r_t^i$ respectively. The definition of $B_{\Ucal'}$ suggests the following consistent definition of $\Ucal'$
\begin{align}
\Ucal'=\{i_g|i \in \Ucal\}\cup\{i_r\}
\end{align}

Let us denote the set of neighbor node of $M$ as $\Pcal$. According to \cite{DioComWou03}, for almost any choice of $A$, the rank of the transfer matrix $T(z)$ and $T_{\Ucal'}(z)$ are
\begin{align*}
\rank(T(z)) = \; &\textrm{max. \# of vertex disjoint paths} \\
& \textrm{from all \emph{input} nodes to }\{M\}\cup\Pcal \\
\rank(T_{\Ucal'}(z)) = \; &\textrm{max. \# of vertex disjoint paths} \\
& \textrm{from }\Ucal'\textrm{ to }\{M\}\cup\Pcal
\end{align*}
For each vertex disjoint path starting from $i_g,i \in \Ncal$, placing $i_g$ with $i_r$ also forms a vertex disjoint path. We can conclude that $\rank(T(z))=\rank(T_{\Ucal'}(z))$. Therefore the sequence $Y_t$ cannot determine a unique sequence of subgradients $\Gtilde_t^\Ncal$.

\bibliography{bibfile}
\bibliographystyle{unsrtnat}
\end{document}